\documentclass[conference]{IEEEtran}

\RequirePackage{ifpdf}

\ifpdf
  \pdfoutput=1\relax
  \usepackage{graphicx}
  \DeclareGraphicsExtensions{.pdf,.png,.jpg,.jpeg}
\else
  \usepackage{graphicx}
  \DeclareGraphicsExtensions{.eps}
\fi

\usepackage{cite}
\usepackage{amsmath,amssymb,amsfonts}
\usepackage{algorithmic}
\usepackage{textcomp}
\usepackage[dvipsnames]{xcolor}

\usepackage{array}
\usepackage{url}
\usepackage[colorlinks=true,linkcolor=red,urlcolor=blue,pdfstartview=FitH,bookmarksopen=true, citecolor=green]{hyperref}

\usepackage{caption}
\usepackage{listings}

\usepackage{etoolbox}
\usepackage{tikz}

\def\BibTeX{{\rm B\kern-.05em{\sc i\kern-.025em b}\kern-.08em T\kern-.1667em\lower.7ex\hbox{E}\kern-.125emX}}

\newrobustcmd*{\mycircle}[1]{\tikz{\filldraw[draw=#1,fill=#1] (0,0) circle [radius=0.1cm];}}
\newrobustcmd*{\mytriangle}[1]{\tikz{\filldraw[draw=#1,fill=#1] (0,0) -- (0.2cm,0) -- (0.1cm,0.2cm);}}

\begin{document}

\title{Computer Vision-aided Atom Tracking in STEM Imaging}

\author{

\IEEEauthorblockN{1\textsuperscript{st} Yawei Hui}
\IEEEauthorblockA{\textit{\small Computer Science and Mathematics Division} \\
\textit{\small Oak Ridge National Laboratory}\\
\small Oak Ridge, TN 37831 \\
\href{mailto:huiy@ornl.gov}{huiy@ornl.gov}
}

\and

\IEEEauthorblockN{2\textsuperscript{nd} Yaohua Liu}
\IEEEauthorblockA{\textit{\small Neutron Scattering Division} \\
\textit{\small Oak Ridge National Laboratory}\\
\small Oak Ridge, TN 37831 \\
\href{mailto:liuyh@ornl.gov}{liuyh@ornl.gov}
}

}

\maketitle

\begin{abstract}
To address the SMC'17 data challenge -- ``Data mining atomically resolved images for material properties'', we first used the classic ``blob detection'' algorithms developed in computer vision to identify all atom centers in each STEM image frame. With the help of nearest neighbor analysis, we then found and labeled every atom center common to all the STEM frames and tracked their movements through the given time interval for both Molybdenum or Selenium atoms.\footnotemark
\end{abstract}


\section{Introduction}
\label{sec:Introduction}

In this response to the data challenges presented by the Smoky Mountain Computational Science and Engineering Conference, we try to answer the specific questions raised in Challenge \#3 which ``is driven by efforts in Scanning Transmission Electron Microscopy to expedite materials data analysis, and generate insight into physics and chemistry of 2D materials irradiated by the electron beam.'' The STEM data sets provided for this challenge are frames of STEM images which record the intensity map of the hexagonal MoSe$_2$ mono-layer with significant defects and dynamic structural re-arrangement. Along the path to find solutions, we considered several algorithms for each question/task. In the following sections, we lay out our best solutions and give a brief discussion at the end about how to improve our result given further research time and resources. All our solutions are implemented in R with certain utility libraries preloaded. Among them there are several critical packages such as ``imager'', ``dbscan'' and ``plot3D'', and their functionalities will be explained according to where they are applied.

\section{Solution}
\label{sec:Solution}

\begin{figure}[!h]
	\begin{center}
		\includegraphics[width=0.41\textwidth]{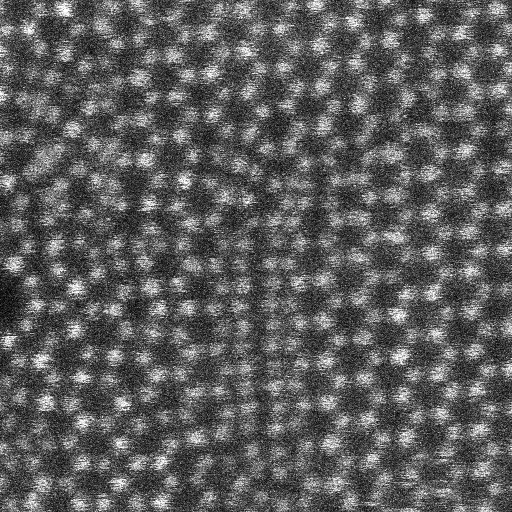} \\
		\vspace{0.1cm}
		\includegraphics[width=0.41\textwidth]{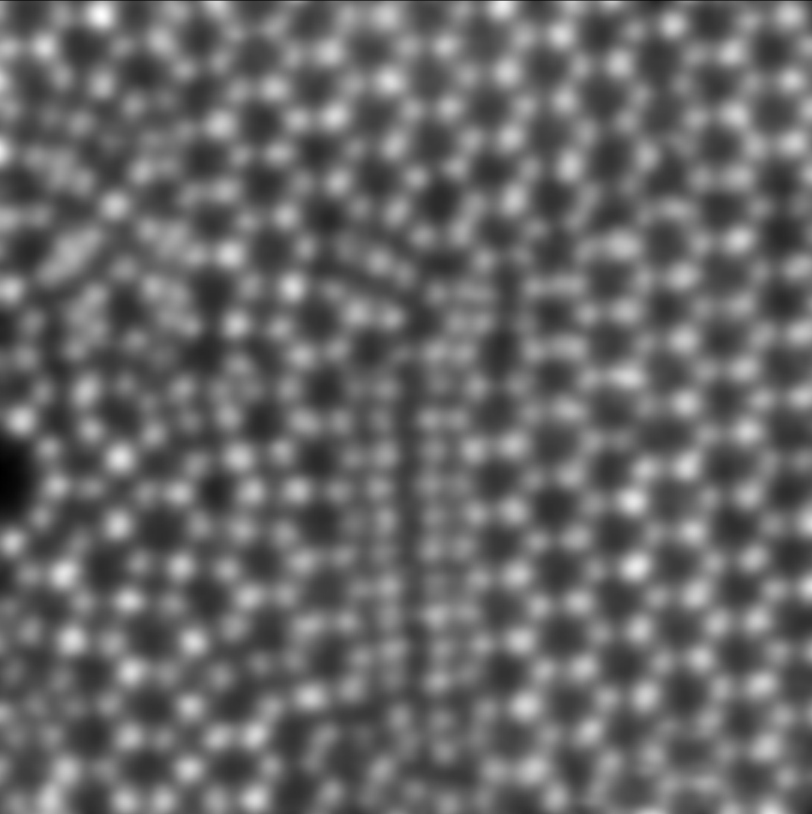}
	\end{center}
	\caption{TOP –- The first frame in the image stack; BOTTOM -– The iso-blurred image as a denoised input for the next step in processing.}
	\label{fig:Frame01}
\end{figure}

\subsection{Question \#1}
\label{sec:Q1}

\textbf{QUESTION:} Identify all atom centers in a single frame, with a robust scalable algorithm capable of identifying all atomic centers in all movie frames. Deliverables: A matrix of $X$, $Y$ positions for every atomic center in Frame 1, and/or A 3D matrix of $X$, $Y$ positions for every atomic center in all frames.

\begin{figure*}[ht]
	\begin{center}
		\parbox{\textwidth}{
			\parbox{0.5\textwidth}{
				\includegraphics[width=0.5\textwidth]{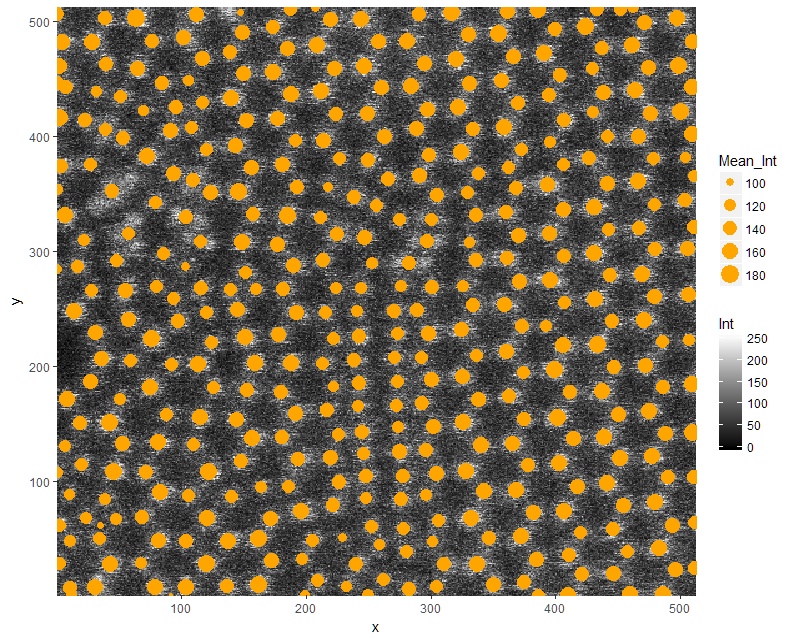}
			}
			\parbox{0.5\textwidth}{
				\includegraphics[width=0.51\textwidth]{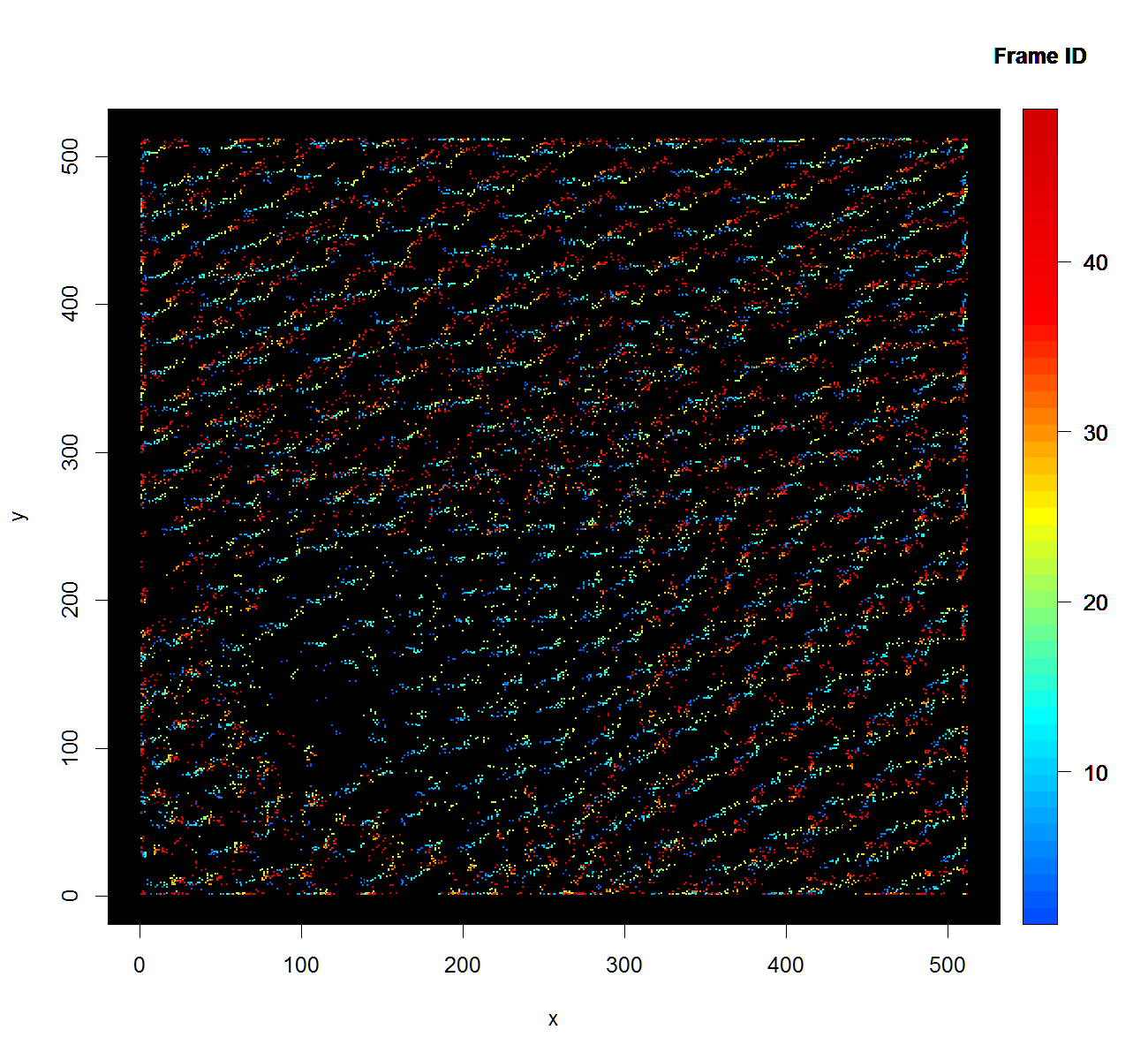}
			}
			\caption{LEFT -- (a) The identified atom centers displayed over the original first frame in the image stack. At each point, the average intensity is calculated from the regional mean within the bright blob; RIGHT -- (b) All identified atoms in each image frame are plotted with color representing the frame ID in which they are recorded. }
			\label{fig:AtomMap}
		}
	\end{center}
\end{figure*}

\footnotetext{VIDEO LINK: \url{https://www.youtube.com/watch?v=YBAbf3g2i64&t=4s}}

\textbf{SOLUTION:} We picked the classic ``blob detection'' algorithms~\cite{b1, b2, b3} developed in computer vision to tackle this problem. Initially, we examined several unsupervised clustering algorithms such as k-means, dbscan, etc., and found that they weren’t working efficiently. Before starting the blob detection, we have applied certain smoothing function on each frame of the image stack for the purpose of denoising. Many methods could be used in this step and we choose the simplest isotropic blurring (a Gaussian profile with scaled standard deviation at 4). An example is shown in~\autoref{fig:Frame01} for both the RAW data and the denoised images ready for the next step of processing. 

In the detection of the blobs (i.e. the brightly contrasted regions representing atoms), the key idea is to use the gradient information at various levels on the intensity map. Ideally, the intensity boundaries between bright atoms and the dark background could be identified with the directional gradients (or simply their amplitudes). In our case, however, we are trying to find the whereabouts of all atoms (meaning the maximum intensities' locations). To do so, we have to calculate not only the first-order derivatives but also the second-orders. By using the determinant of the Hessian matrix at each pixel point as a weighting factor in determining the maximum position within a blob, we could easily pick up the maxima of blobs' intensities within certain intensity ranges. However, if we are aiming to automatically pick up all blobs whose intensities span a significant dynamic range, we have to resort to the well-established ``scale-space'' theory for blob detection in computer vision~\cite{b4}. With the convenient function ``hessdet()'' in the R package ``imager'', we calculated the Hessian determinant map at various scales and picked the maximum determinants for each pixel. With such a compiled map, the intensity regions containing blobs are identified with 60 percentiles of determinants on the map and the average position and intensity within each region is recorded for references as detected atom centers.

As an example of the solution, we show the identified atom centers overlapped with the original image in~\autoref{fig:AtomMap}\textcolor{red}{(a)}. All the possible centers of atoms appeared in all 49 frames of image are detected and the result is summarized in~\autoref{fig:AtomMap}\textcolor{red}{(b)} (a scatterplot drawn with ``scatter2d()'' function in the R package ``plot3D''). Limited by the resolution of the figure, it's not intuitive to see the overall tracks of the identified atoms moving with time. In the supplemented materials, the animated GIF ``Atom\_Position'' gives a much better dynamic view of the movement of all the identified atoms.

\subsection{Question \#2}
\label{sec:Q2}

\textbf{QUESTION:} Identify and label every atomic center common to all frames. Deliverables: A matrix of $X$, $Y$ locations for atomic centers that can be found in all 49 frames. Each center needs to carry a unique identifier allowing the same atomic to be referenced in any of the 49 frames.

\begin{figure}[!h]
	\begin{center}
		\includegraphics[width=0.5\textwidth]{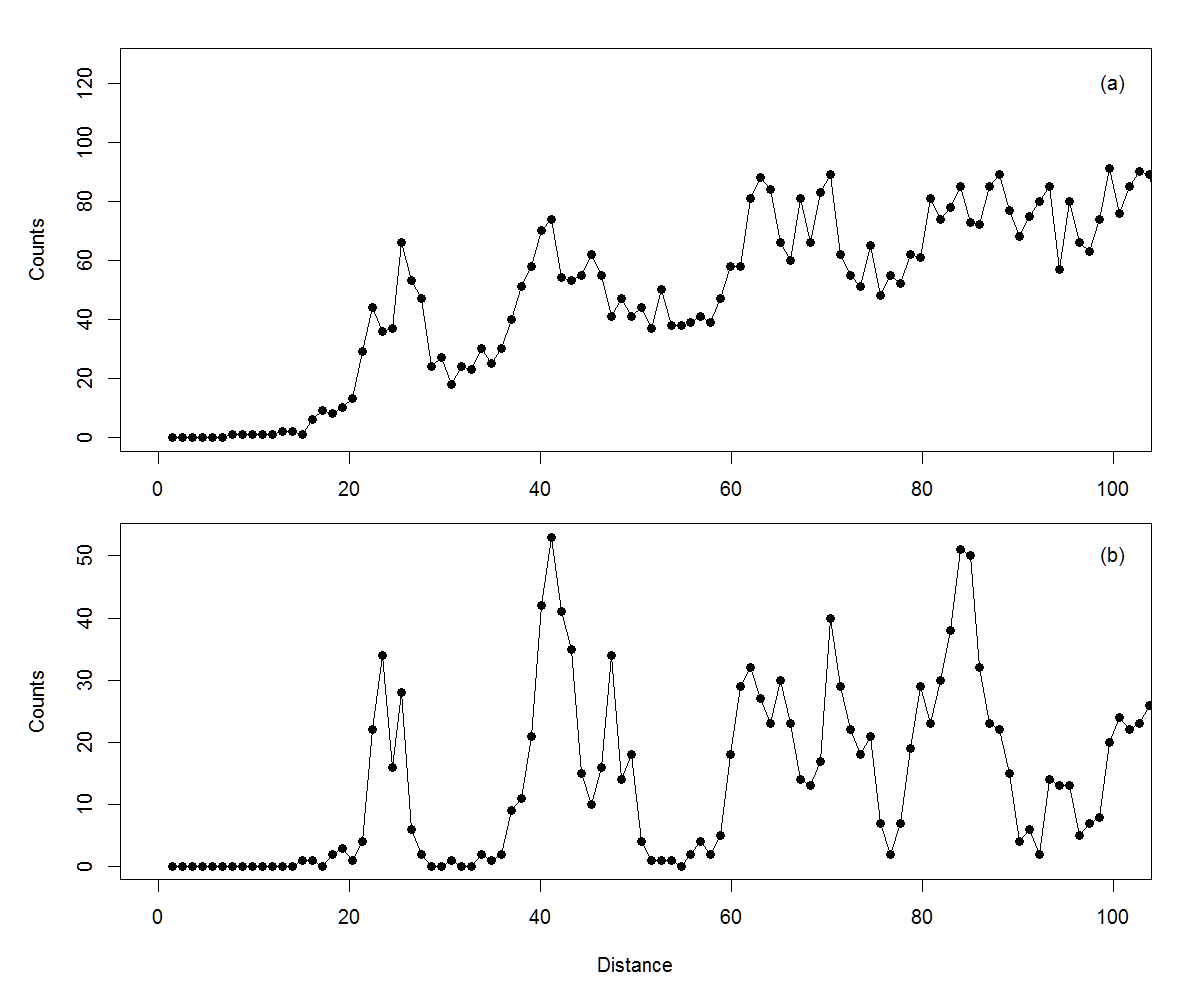}
	\end{center}
	\caption{(a) The histogram of distances among all paired positions calculated in all frames. (b) Instead of all positions in all frames, this histogram shows the distances among paired positions in regular lattice region in x = [300, 512] and y = [0, 300].}
	\label{fig:AtomNNDist}
\end{figure}

\textbf{SOLUTION:} As the starting positions for trackable atoms in all frames, the identified atom centers in frame \#1 are taken for granted. Tracking their movements is processed as we try to associate as accurately as possible the identified centers between two adjacent frames. Given the identified atom centers in frame \#$i$, we calculate distances from a single atom center ($P$) identified in frame \#($i$+1) to all centers in frame \#$i$. We then pick the minimum distance between $P$ and a center $Q$ in frame \#$i$, and compare it with a characteristic distance (hereafter, $R_0$) which represents half of the length of the Mo-Se bound in the crystal lattice. If the minimum distance is shorter than $R_0$, we consider that the atom at $P$ is moved from the center at $Q$ in the previous frame. Otherwise, we drop the atom at $P$ from the list of trackable atoms. One thing to note is how we decide $R_0$. In~\autoref{sec:Q1}, after all the centers in all frames are determined, we calculate distances of all paired positions in a single frame. In~\autoref{fig:AtomNNDist}, two histograms show the distribution of the calculated distances with~\autoref{fig:AtomNNDist}\textcolor{red}{(a)} including all paired positions in all frames, and~\autoref{fig:AtomNNDist}\textcolor{red}{(b)} including only the region showing regular lattice structures in x = [300, 512] and y = [0, 300]. It is clear to see that the ``nearest neighbor'' distances lie in the range of 20 to 30 pixels. For the atom tracking, we choose $R_0=15$, which is half of the maximum ``nearest neighbor'' distance.

With $R_0=15$, we can track in all frames 124 atoms in total. In~\autoref{fig:AtomTrack}, we show the centers of all trackable atoms in all frames and in the supplemented materials, the animated GIF ``Atom\_Track\_ UnIdentified'' gives a much better dynamic view of the movement of all trackable atoms.

\begin{figure}[!h]
	\begin{center}
		\includegraphics[width=0.5\textwidth]{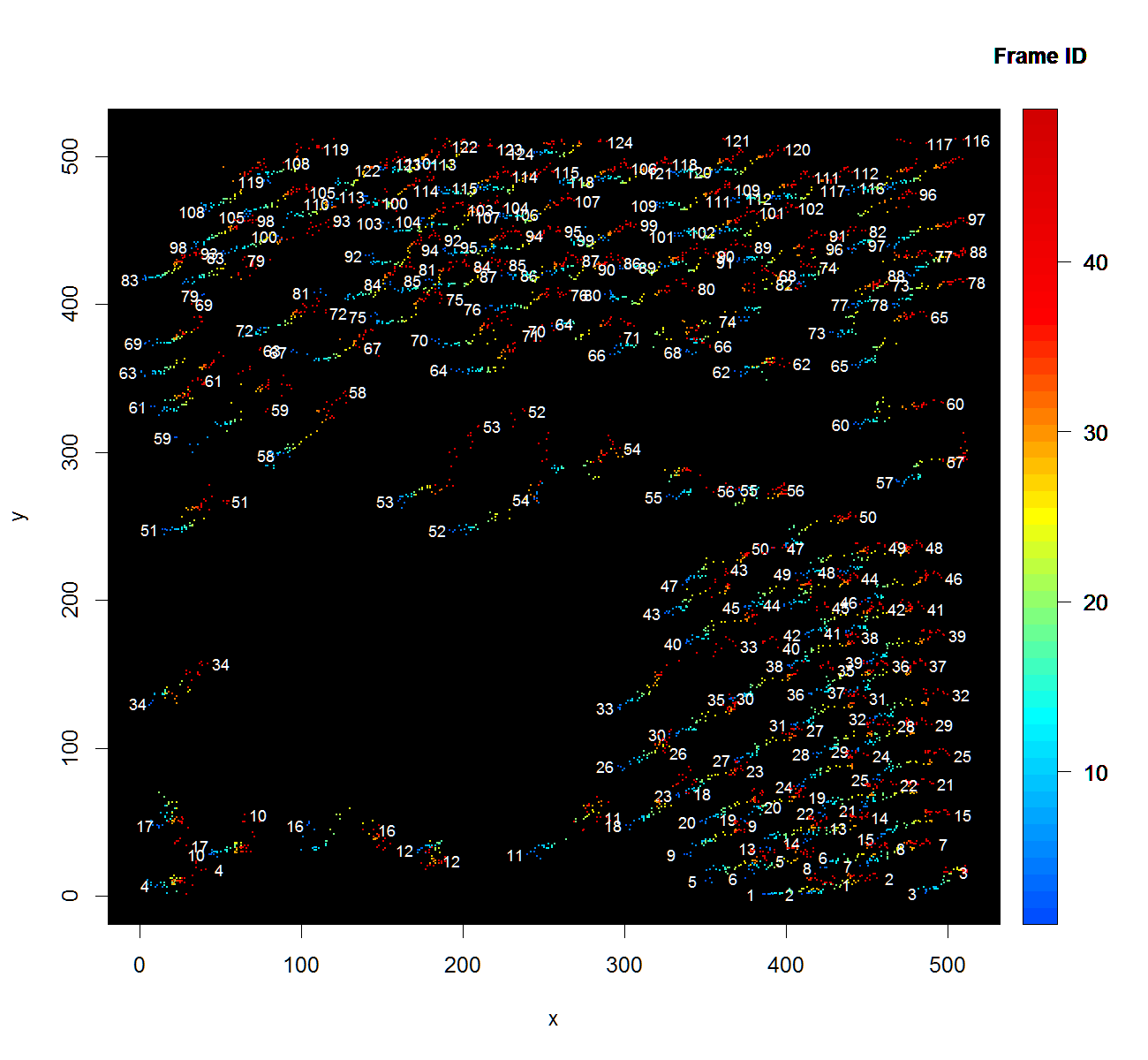}
	\end{center}
	\caption{All identified atoms which could be tracked in all image frames are plotted with color representing the frame ID in which they are recorded. The unique atom ID is marked with two numbers at the beginning and end of the track (e.g. atom ID 34).}
	\label{fig:AtomTrack}
\end{figure}

\subsection{Question \#3}
\label{sec:Q3}

\textbf{QUESTION:} Create a vector map of atomic motion for each of the uniquely identified atomic centers common to all 49 frames. Create a graphic that captures trajectories for all atomic centers common to the movie. Deliverable: A vector array of $X$, $Y$ positions for each uniquely identified atomic center throughout all the movie frames. A graphic illustrating full 49 frame trajectories for one, some, or all atomic centers.

\textbf{SOLUTION:} This question is a natural extension of question \#2 and its solution has been given in the last section (in the supplemented GIF).

\subsection{Question \#4}
\label{sec:Q4}

\textbf{QUESTION:} Molybdenum and Selenium have different intensities in the image. Selenium atoms are ~8\% brighter than the Molybdenum. Furthermore, in an ideal crystal, the locations of Mo and Se atoms in any hexagon in the image are rigidly defined in what can be defined as ''upwards and downwards facing triangles'' of Mo and Se atoms. In this challenge we are interested in identity of the labeled atoms common to all frames. Identity can be determined from intensity, crystallographic orientation, or both. Deliverable: A vector array of $X$, $Y$ positions for each uniquely identified atomic center throughout all the movie frames, labeled as either Mo, Se, or Unknown. A graphic illustrating full 49 frame trajectories for one, some, or all atomic centers for either Mo, or Se species.

\begin{figure}[!h]
	\begin{center}
		\includegraphics[width=0.5\textwidth]{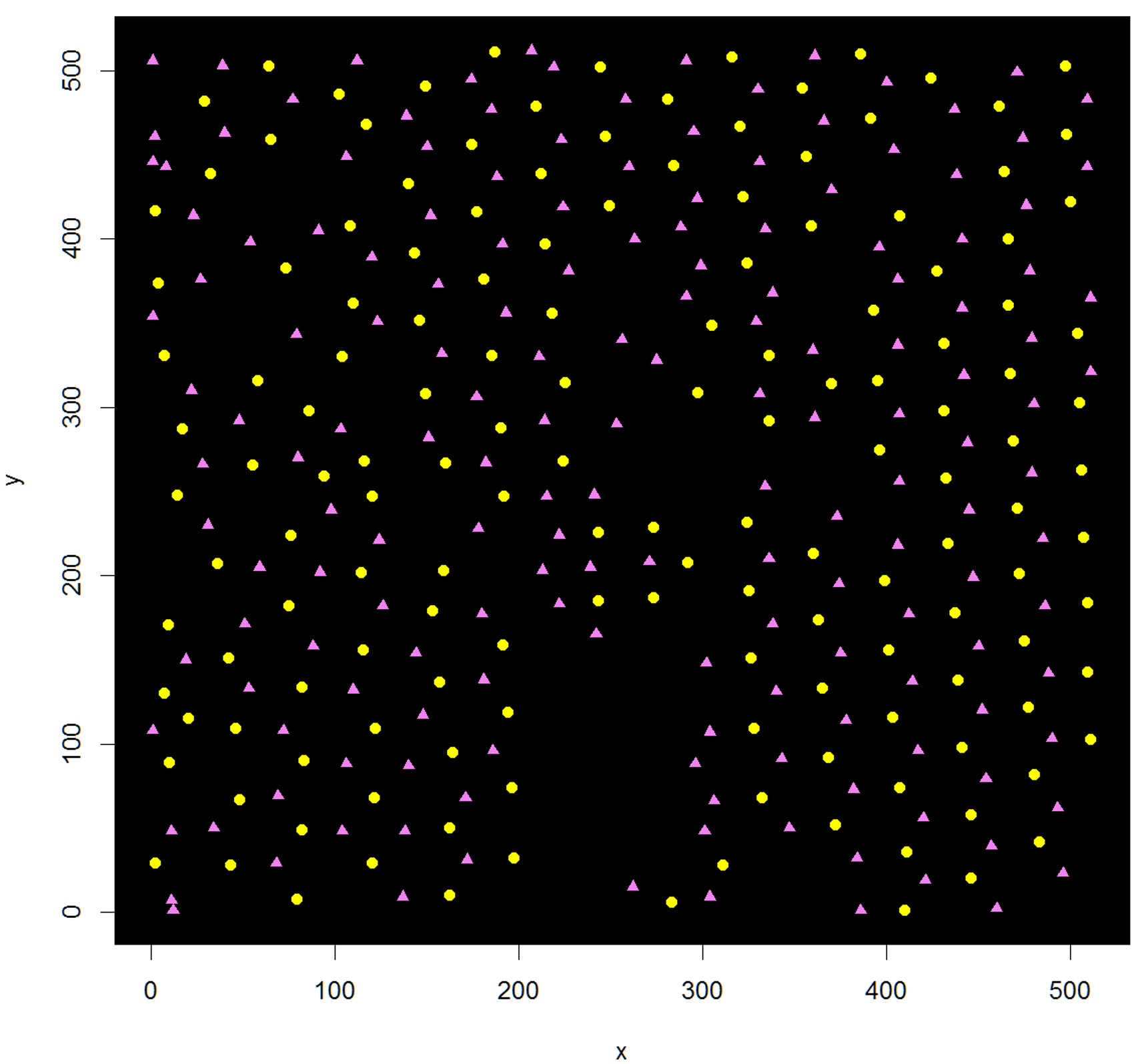}
	\end{center}
	\caption{All identified positions for two different atoms in the first image frame. Yellow dots represent Se atoms and violet triangles Mo atoms, respectively.}
	\label{fig:AtomID}
\end{figure}

\textbf{SOLUTION:} To solve this problem, we started with the relatively reliable crystallographic orientation for Mo and Se atoms in the crystal lattice. In each image frame, the detected atom centers are fed into a R function ``frNN()'' (in package ``dbscan'') which finds the nearest neighbors for each center within a fixed radius. Referring to~\autoref{sec:Q2}, we choose the fixed radius to be the maximum distance of the first nearest neighbor (30 pixels). At each center, we check if there are 2, 3 or 4 neighbors located within an annulus of inner and outer radii 20 and 30, respectively, and by doing so, we guarantee that only first nearest neighbors are selected. In case of having 2 neighbors, we consider they may belong to a lattice cell which happens to locate at the edge of either the image or some void areas. The majority of the detected centers will have 3 neighbors reflecting the regular Mo-Se crystallographic structure. We also consider 4 neighbors simply because we try to make the distinction between the regular structures (hexagon) and the irregular ones (such as shown in the lower middle region of the first image). Once the surrounding environment of a center satisfies these requirements, we compare the intensity at the center with the mean intensity of all its neighbors. If the intensity is larger comparing to its neighbors' mean, we make a vote at this center for a Se atom, otherwise a Mo atom. This procedure is carried out at every center in a single image and at the end, every center will have two records of votes for both Se and Mo atoms. The last step is to sum up all votes for Se and Mo respectively and choose the atom type with more votes than the other. As an example of the final identification, we show in~\autoref{fig:AtomID} the identifiable centers for both Mo and Se atoms and all un-identifiable positions are not marked out.

To determine the type of every trackable atom found in~\autoref{sec:Q2} and~\autoref{sec:Q3}, we consider the atom's identification along all 49 image frames. By counting how many times it is categorized as one type and comparing among the total counts of three available categories – Un-identified, Mo and Se – we pick the atom type which has the most counts. In the supplemented materials (GIF animation ``Atom\_Track''), we show the dynamic tracks of these three atom types in all image frames. The legends used in the animation are that ``X'' represents Un-identified, \mycircle{yellow} Molybdenum, and \mytriangle{pink} Selenium, with relatively sized symbols representing the mean intensities.

\section{Discussion}
\label{sec:Discussion}

\subsection{Image Denoising}
\label{sec:ImageDenoising}

We used the isotropic blurring technique for the image denoising before the blob detection in~\autoref{sec:Q1}. Better solutions exist, for example, by using ptychography processing which is based on Principle Component Analysis (PCA) (tested without showing results due to page limitation). However, if it's going to be applied to a real-time analysis for atom tracking and identification, PCAs will take significant portion of the processing time. For the purpose of validation of idea, the iso-blurring algorithm is sufficiently good for our proposed atom center detection approach.

\subsection{Nested Blob Detection}
\label{sec:NestBlob}

In many images, there are cases that two detected atom centers are located closely enough so that they are taken by the algorithm as a single blob. To further distinguish and separate them apart, one way is to run the same Hessian-based algorithm for each detected region. Deliberate logics should be applied, if taken this nested operation, to guarantee the consistency, though.

\subsection{Global Optimization}
\label{sec:GlobalOptimization}

The most significant improvement in our solutions we'd like to make, if time and resources allowed, is to iterate around the atom center tracking and identification consistently. In our implementation, we do the center identification solely dependent on the regular Mo-Se crystallographic structure after the blob detection. Following that, the movement tracking takes into account only identity information based on changing positions without consistently considering identity information based on atomic types. In a global optimization, we envision that certain voting weights should be granted to the two different identities and a controlled iteration built around the procedures in~\autoref{sec:Q2} and \autoref{sec:Q4} would reach a point where those two identities are consistent with each other eventually.

\section*{Acknowledgment}
\label{sec:Acknowledgment}

This manuscript has been authored by UT-Battelle, LLC under Contract No. DE-XXXX-00ORXXXXX with the U.S. Department of Energy. The United States Government retains and the publisher, by accepting the article for publication, acknowledges that the United States Government retains a non-exclusive, paid-up, irrevocable, world-wide license to publish or reproduce the published form of this manuscript, or allow others to do so, for United States Government purposes. The Department of Energy will provide public access to these results of federally sponsored research in accordance with the DOE Public Access Plan (\url{http://energy.gov/downloads/doe-public-access-plan}).

We thank Dr. Stephen Jesse and Dr. Raymond Unocic at the Center for Nanophase Materials Science (CNMS) for insightful discussions on related topics.

\end{document}